\pdfoutput=1
\documentclass{article}

\usepackage{arxiv}

\PassOptionsToPackage{numbers}{natbib}

\usepackage[utf8]{inputenc} 
\usepackage[T1]{fontenc}    
\usepackage{url}            
\usepackage{booktabs}       
\usepackage{amsmath}
\usepackage{amsfonts}       
\usepackage{nicefrac}       
\usepackage{microtype}      
\usepackage{lipsum}
\usepackage[caption = false]{subfig}
\usepackage[linesnumbered,lined,boxed]{algorithm2e}
\usepackage{multicol}
\usepackage{multirow}
\usepackage{graphicx}

\newtheorem{definition}{Definition}

\title{Deep Neural Networks Predicting Oil Movement in a Development Unit}

\author{
  Pavel Temirchev\textsuperscript{a}\thanks{Corresponding author.} \\
  \texttt{P.Temirchev@skoltech.ru} \\
  \And
  Maxim Simonov\textsuperscript{b} \\
  \texttt{Simonov.MV@gazpromneft-ntc.ru} \\
  \And
  Ruslan Kostoev\textsuperscript{a} \\
  \texttt{R.Kostoev@skoltech.ru} \\
  \AND
  Evgeny Burnaev\textsuperscript{a} \\
  \texttt{E.Burnaev@skoltech.ru} \\
  \And
  Ivan Oseledets\textsuperscript{a} \\
  \texttt{I.Oseledets@skoltech.ru} \\
  \And
  Alexey Akhmetov\textsuperscript{b} \\
  \texttt{Akhmetov.AV@gazpromneft-ntc.ru} \\
  \AND
  Andrey Margarit\textsuperscript{b} \\
  \texttt{Margarit.AS@gazpromneft-ntc.ru} \\
  \And
  Alexander Sitnikov\textsuperscript{b} \\
  \texttt{Sitnikov.AN@gazpromneft-ntc.ru} \\
  \And
  Dmitry Koroteev\textsuperscript{a} \\
  \texttt{D.Koroteev@skoltech.ru} \\
}

\begin{document}
\maketitle

\bigskip
\textsuperscript{\bf a} Skolkovo Institute of Science and Technology, \\Nobelya Ulitsa 3, Moscow, Russia 121205
\\
\\
\textsuperscript{\bf b} Gazpromneft Science and Technology Centre, \\75-79 liter D Moika River emb., Saint Petersburg, Russia 190000
\\
\bigskip

\begin{abstract}

We present a novel technique for assessing the dynamics of multiphase fluid flow in the oil reservoir. We demonstrate an efficient workflow for handling the 3D reservoir simulation data in a way which is orders of magnitude faster than the conventional routine. The workflow (we call it ``Metamodel'') is based on a projection of the system dynamics into a latent variable space, using Variational Autoencoder model, where Recurrent Neural Network predicts the dynamics. We show that being trained on multiple results of the conventional reservoir modelling, the Metamodel does not compromise the accuracy of the reservoir dynamics reconstruction in a significant way. It allows forecasting not only the flow rates from the wells, but also the dynamics of pressure and fluid saturations within the reservoir. The results open a new perspective in the optimization of oilfield development as the scenario screening could be accelerated sufficiently.

\end{abstract}


\section{List of Abbreviations}

\begin{tabular}{|c|l|}
\hline
FDHS & Finite-Difference Hydrodynamical Simulator\\ 
\hline
ROM & Reduced Order Modelling\\ 
\hline
CRM & Capacitance-Resistance Model\\ 
\hline
PDE & Partial Differential Equation\\ 
\hline
GRU & Gated Recurrent Unit\\
\hline
(n)MDP & (non-)Markovian Decision Process\\
\hline
ML & Machine Learning\\ 
\hline
PCA & Principal Components Analysis\\ 
\hline
SVD & Singular Value Decomposition\\ 
\hline
VAE & (Variational) Autoencoder\\ 
\hline
ELBO & Evidence Lower Bound Objective\\ 
\hline
LReLU & Leaky Rectified Linear Unit\\ 
\hline
LR & Linear Regression\\ 
\hline
FCNN & Fully-Connected Neural Network\\ 
\hline
RNN & Recurrent Neural Network\\ 
\hline
NN & Neural Network\\ 
\hline
DDF & Data-Driven Forecasting\\ 
\hline
\end{tabular}

\section{Introduction}

Computer 3D modelling of oil flows through a porous medium is the most frequently used tool for an oil field development optimization and prediction of unknown reservoir properties by history-matching procedure \cite{history-matching-1990}. It is essential to have an accurate and fast enough model of oil flows to solve both problems.

The common modelling approach is to use a finite-difference hydrodynamical simulation (FDHS) \cite{ertekin-bars-2001}. FDHS uses computations on a discrete computational grid, and its accuracy is strongly dependent on a resolution of the grid. Being accurate, computations on fine grids may consume weeks to be performed for a big oil field. It is almost impossible to run iterative optimisation and history-matching methods using this kind of models in sufficient time.

There are many techniques aimed to speed up the simulation process: starting from simple downscaling and local grid refinement and ending with data-driven Reduced Order Modelling (ROM) approaches. Capacitance-Resistance Models (CRM)~\cite{connectivitycrm-2005, kovalcrm-2015, fullycoupledcrm-2014}, for example, tries to fit nonlinear autoregressive model able to forecast fluid production rates. While CRM retains the physicality of the process, other models train autoregressive production model in a fully data-driven manner~\cite{insim-2015, rpm-2016, shm-2layer-2015}. On the other hand, Galerkin projection methods are aimed to find the linearly reduced approximation of Partial Differential Equations (PDE) solved by FDHS~\cite{drrnn-2018, mor-in-fd, srom, pwl-mor}. The overall idea of the approach is to construct projection bases using Proper Orthogonal Decomposition (POD).  For nonlinear PDEs straight Galerkin projection is computationally inefficient due to the need of computing fully-dimensional Jacobian matrix at each timestep. Some approaches tries to overcome this issue via additional approximation layer, e.g. Empirical Interpolation~\cite{eim} and Discrete Empirical Interpolation~\cite{deim} methods.

Petroleum society requires a fast and accurate model for three-phase flow simulation on a 3-dimensional computational grid. And all the discussed approaches cannot solve the task properly: either they are not able to produce solutions for spatial fields (pore pressure, saturation, etc.) or they can work only 
under nonrealistic 
greatly simplified settings.

In this work, we provided a methodology for a fast data-driven 3D reservoir modelling, based on machine learning techniques and aimed to approximate FDHS solutions on a subset of possible simulation options. We use ``Metamodel'' terminology \cite{GTApprox2016} to describe a purely data-driven 3D reservoir model approximating the solution given by a base model (e.g. FDHS), and we provide a detailed definition in Section~\ref{metamodelling-sec}.

We concentrated our attention on modelling so-called development units - symmetric subelements of a whole oilfield. We discussed scenarios with water injection and without injection at all. The approach can be straightforwardly used to train a metamodel for different injection fluids though this theme is out of the scope of this paper.

The idea of forecasting for just a development unit can be used to construct a metamodel of a whole oilfield via division it into several independent units. On the other hand, fully-convolutional neural networks can represent a way to scale the proposed approach to a whole oilfield. Experiments with whole and real cases are retained for future work.

Proposed metamodelling approach is based on the idea of forecasting in latent variable space. The same idea was used in~\cite{e2c-2015, rce-2017} to approximate the dynamics of an environment with high-dimensional state space for model-based Reinforcement Learning. Authors propose to use Variational Autoencoders~\cite{vae-2013} for the model order reduction and we try it in our framework.

Our major contributions are:

\begin{itemize}
    \item The metamodelling framework based on forecasting in latent variable space. 
    \item Metamodel of a field development unit composed of Convolutional Variational Autoencoder for dimensionality reduction and GRU Recurrent Neural Network~\cite{gru-2014} for the latent dynamics prediction. The metamodel is able to predict both pressure-saturation distributions (Section~\ref{latent-dynamics-sec}) and production rates from wells (Section~\ref{well-prodrates-sec}).
    \item Comparative analysis of different metamodels in the sense of FDHS approximation accuracy, see  Section~\ref{experiments-sec}. 
\end{itemize}

\section{Metamodelling of Reservoir Flows}\label{metamodelling-sec}
We mentioned that our approach is developed to work in a subset of possible simulation options. The overall operational envelope of the presented model has two major restrictions:

\begin{enumerate}
    \item Geometry of the computational domain is restricted to a field development unit,
    \item Reservoirs are homogeneous in horizontal dimensions.
\end{enumerate}

\begin{figure}[ht!]
    \centering
    \subfloat[Staggered line drive]{\label{ascheme:staggered}\includegraphics[width=.3\linewidth]{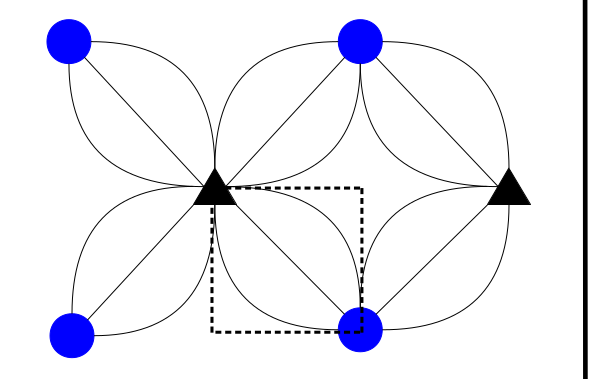}}
    \subfloat[Direct line drive]{\label{ascheme:direct}\includegraphics[width=.3\linewidth]{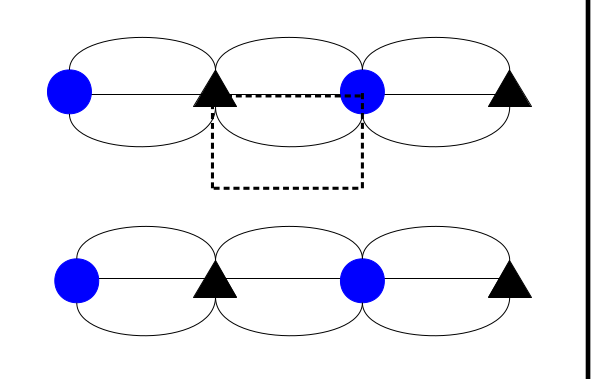}}
    \subfloat[Without injection]{\label{ascheme:no_inj}\includegraphics[width=.3\linewidth]{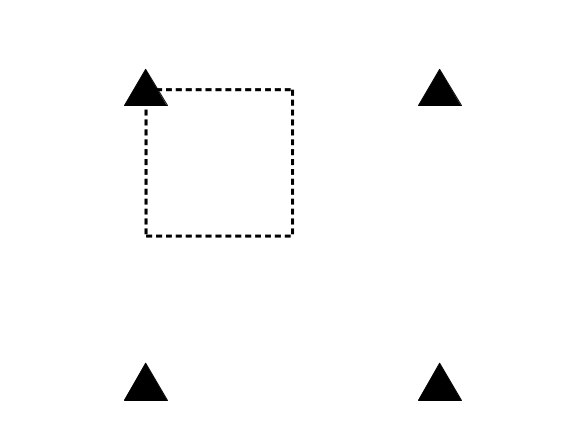}}
    \caption{Investigated well allocation schemes with circles used to denote injection wells and triangles to denote production wells. Dashed lines segregate 1/4-th of development units.}
    \label{ascheme}
\end{figure}

It may be noticed that if wells are allocated on a regular grid and the reservoir is homogeneous in horizontal dimensions, then the dynamics will be the same for all elements of the grid. In petroleum engineering, such an element is usually referred to as a development unit. We investigate three common well allocation schemes (fig. \ref{ascheme}):

\begin{itemize}
    \item Staggered line drive (fig. \ref{ascheme:staggered}),
    \item Direct line drive (fig. \ref{ascheme:direct}),
    \item Without injection (fig. \ref{ascheme:no_inj}).
\end{itemize}

A model that forecasts dynamics for a development unit (actually for $1/4$-th of it) may be directly applied for symmetric reservoirs. Though real reservoirs frequently are neither strictly symmetric nor horizontally homogeneous, they may be discretised into development units of required form. The ensembling of units produced by such a discretisation was left for the future work. Throughout the rest of the paper, we consider forecasting dynamics and production rates for 1/4-th of a development unit with one production well and, possibly, one injection well.

We will think of a reservoir dynamics as a non-Markovian Decision Process (nMDP) with a time discretization step $\Delta t$. The nMPD is defined using 5-tuple $(\mathcal{F}, \mathcal{U}, \mathcal{M}, F, R)$, where:

\begin{itemize}
    \item {$\mathcal{F}$ is a set of reservoir states $f_t \in \mathcal{F}$ representing distributions of pore volume pressure $p_t$ and oil, water and gas saturations $s_t^o, s_t^w, s_t^g$ within 3D coordinate grid at a fixed moment of time $t$. Since $s_t^o(x, y, z) + s_t^w(x, y, z) + s_t^g(x, y, z) = 1$ we may drop one of the saturations from the reservoir state and compute it explicitly when needed (e.g. gas saturation).} 
    \item {$\mathcal{U}$ is a set of control variables $u_t \in \mathcal{U}$ representing production bottomhole pressure and water injection rate (if injection is presented).}
    \item {$\mathcal{M}$ is a set of variables $m \in \mathcal{M}$ defining reservoir initial conditions, rock and fluids properties. This variable is invariant of time and is fixed for a concrete development unit. We refer to these variables as a metadata of a development unit.}
    \item {$F$ is a transition function that maps applied control sequence $u_{0:t}$, all previous states $f_{0:t}$ and metadata of development unit $m$ into a following reservoir state $f_{t+1}$, such that $f_{t+1} = F(f_{0:t}, u_{0:t}, m)$}
    \item {$R$ is a production rates function, such that $r_t = R(f_{0:t}, u_{0:t}, m)$, where $r_t$ is a vector of oil, water and gas production rates at time $t$. Production rates may be thought as a kind of analogy to reward in classical MPDs.}
\end{itemize}

There are models able to restore the transition function $F$, and production rates function $R$ by solving corresponding partial differential equations (PDEs). We call them Base Models. Commonly, Base Models solve PDEs using finite-difference or finite-volume schemes on a discretised grid and differ from each other in used PDE's modifications and numerical integration schemes. Further, we will concentrate on so-called finite-difference hydrodynamical simulators (FDHS), which solve three-phase fluid flow equations with gas liberation/dissolving from/into the oil options. Though, our approach can be straightforwardly generalised for other types of Base Models.

Thinking of FDHS as of an ideal flow model (in the sense of accuracy) we may want to speed up it by substituting it with a metamodel. To proceed, we need a formal definition of a reservoir metamodel (see also \cite{GTApprox2016}):

\begin{definition}
Reservoir metamodel is a data-driven reservoir model aimed to approximate the transition function $F$ and production rates function $R$ represented by a Base Model.
\end{definition}

Here and further we will assume the approximation quality to be measured in the sense of L2-norm. If we denote approximated transition and production functions by $\hat F$ and $\hat R$ respectively then the approximation problem may be formulated as follows:

\begin{equation}
    \mathbf{E}_{m, u} \Big[ \frac{1}{T} \sum_{t=0}^T ||\hat F(\hat f_{0:t}, u_{0:t}, m) - F( f_{0:t}, u_{0:t}, m)||^2_2 \Big] \rightarrow \min_{\hat F},
\end{equation}
\begin{equation}
    \mathbf{E}_{m, u} \Big[ \frac{1}{T} \sum_{t=0}^T ||\hat R(\hat f_{0:t}, u_{0:t}, m) - R( f_{0:t}, u_{0:t}, m)||^2_2 \Big]  \rightarrow \min_{\hat R},
\end{equation}
where the sequence of approximated reservoir states $\hat f_{0:t}$ is obtained from the approximated transition function $\hat F$, and the initial states are computed as follows:

\begin{equation}\label{initial_states}
    f_0 = F^{init}(m), \qquad \hat f_0 = \hat F^{init}(m).
\end{equation}
 
In the following sections, we will discuss the training procedure of a data-driven reservoir metamodel starting from a description of the training set.

\section{Training Data}\label{training-set-sec}

In order to construct a metamodel we use supervised machine learning techniques, requiring a training set of examples of the Base Model dynamics. Training set $\mathbf{D}$ is composed of scenarios (episodes), where each scenario represents one run of FDHS
and contains:

\begin{enumerate}
    \item\label{var-user-defined-enum} {Input variables:
    \begin{itemize}
        \item Scenario metadata in compressed vector form: $m \in \mathbf{R}^{61}$ (for more details see \ref{metadata-appendix}). Metadata is fixed throughout the scenario.
        \item Sequence of control variables $u_{0:T}$, $u_t \in \mathbf{R}^2$ consists of production bottomhole pressure and water injection rate at time $t$.
    \end{itemize}
    }
    \item\label{var-computed-enum} {Output variables (FDHS solutions):
    \begin{itemize}
        \item Sequence of reservoir states $f_{0:T}$. Each reservoir state $f_t \in \mathbf{R}^{3 \times n_x \times n_y \times n_z}$ is a tensor containing values of pore pressure, oil and water saturation for all cells of computational grid at time $t$, where $n_x$, $n_y$, $n_z$ are numbers of cells in corresponding dimensions.
        \item Sequence of oil, water and gas production rates $r_{0:T}$, $r_t \in \mathbf{R}^3$.
    \end{itemize}
    }
\end{enumerate}

The output variables (\ref{var-computed-enum}) were obtained from multiple FDHS runs on randomly generated input variables (\ref{var-user-defined-enum}). The generative model of input variables is based on the analysis of real laboratory data and aimed to produce realistic samples (see \ref{gen-model-appendix} for details).

Each scenario has its own time horizon $T$. FDHS turns off if the oil production rate falls below the limit of 3 stb/day, with an upper bound: $T < 30$ years. 

\section{Latent Variable Reservoir Dynamics}\label{latent-dynamics-sec}

Direct application of basic ML techniques to the approximation of functions $F$ and $R$ is complicated by the so-called curse of dimensionality. Even for simplified Markovian version of the task (when $f_{t+1} = F(f_t, u_t, m)$), the basic linear regression model will require learning of more then $(3 n_x n_y n_z)^2$ parameters, which will lead to the overfitting of the metamodel for any realistic number of available training examples --- the metamodel will not be able to reproduce training accuracy on scenarios not included in the training set. Moreover, such a model will not be efficient both in terms of  computational speed and  memory consumption.

One way to reduce the problem complexity is to find a bijective mapping (dimensionality reduction) $G: \mathcal{F} \rightarrow \mathcal{Z}$, where $\mathcal{Z}$ is a low-dimensional latent variable space, and to define the dynamics in this latent variable space:

\begin{equation}
    z_{t+1} = F^{latent}( z_{0:t}, u_{0:t}, m).    
\end{equation}

\begin{figure}
    \centering
    \includegraphics[width=.6\linewidth]{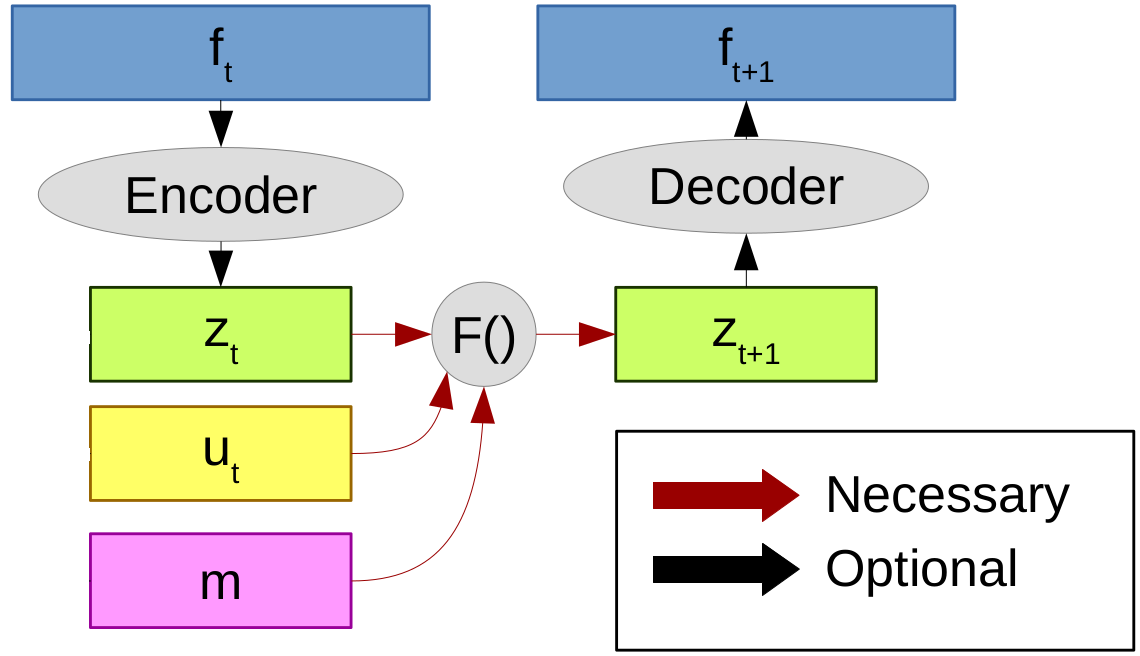}
    \caption{Graphical representation of the latent variable reservoir dynamics workflow, cropped to one timestep.}
    \label{latent-dynamics-scheme}
\end{figure}

The graphical representation of this latent variable reservoir dynamics model is shown in Figure \ref{latent-dynamics-scheme}. 

Full-order reservoir states $f_t$ might be recovered as $f_t = G^{-1}(z_t)$. It is not required to recover $f_t$ at each timestep -- all the dynamics can be computed in the latent variable space with the need to use inverse mapping $G^{-1}$ only for the timesteps of interest.

In the following subsections, we show some practical models for both dimensionality reduction function $G$ and latent dynamics function $F^{latent}$. Different combinations of these models will result in different reservoir metamodels.

\subsection{Search for a mapping $G$ for the dimensionality reduction}\label{autoencoding-subsec}

From an ML point of view, dimensionality reduction problem may be solved by fitting the so-called autoencoding model. We will use autoencoders to represent the true bijective mapping $G$ and its inverse $G^{-1}$ with parametrized approximations $G_\theta$ (encoder) and $G^{-1}_\theta$ (decoder).

Autoencoder passes the input $f_t$ through a low-dimensional bottleneck via subsequent application of encoding and decoding functions, and is trained to minimise some deviation $\delta$ between autoencoder's input and output:

\begin{equation}
    \mathbf{E}_{f_{0:T} \sim \mathbf{D}} \Big[ \sum_{t=0}^T \delta(f_t, G^{-1}_\theta(G_\theta(f_t)) \Big] \rightarrow \min_\theta.
    \label{qq}
\end{equation}

Throughout this work, we analysed two distinct autoencoding models: 
\begin{itemize}
    \item Principal Component Analysis,
    \item Convolutional Variational Autoencoder.
\end{itemize}

\textit{Principal Components Analysis (PCA)} is a linear dimensionality reduction model and is based on Singular Value Decomposition (SVD) of the matrix $X$ composed of all tensors $f_t \in \mathbf{D}$ flattened into vectors:

\begin{equation}
    X = U\Sigma V^T.
\end{equation}

$d_z$ most significant singular values are retained to form truncated matrix of right singular vectors $\tilde V \in \mathbf{R}^{(3n_xn_yn_z) \times d_z}$. $G_\theta(f_t) = \tilde V^T f_t $ is now can be used as an encoder and $G_\theta^{-1}(z_t) = \tilde V z_t$ -- as a decoder.

PCA provides the best approximation accuracy throughout all linear models with fixed latent space dimensionality $d_z$ in the sense of Frobenius norm $\delta(f, f') = ||f -  f'||^2_2$. 

However, PCA has two significant disadvantages: it is linear, and it ignores the spatial structure of the tensor $f_t$ due to the flattening procedure and fully-connected structure. 

\textit{Convolutional Variational Autoencoder} combines the application of Variational Autoencoding framework (VAE)~\cite{vae-2013} with convolutional neural network modules.

Variational Autoencoder is a probabilistic model that tries to approximate not just a point estimate of $z_t$, but rather the posterior distribution over it with probability density function $q_\theta(z_t|f_t)$. The task is solved by maximization of Evidence Lower Bound Objective (ELBO) proposed in the original paper. 

The parametric form of the conditional distribution $q_\theta(z_t|f_t)$ can be chosen to be Gaussian, and in this case, the ELBO maximization is closely related to the minimization of L2-norm of deviation in \eqref{qq}, so we stick with the Gaussian approach.


Due to the deterministic structure of latent dynamics models described below, we must eliminate the stochasticity of VAE. We use the following expressions for encoder and decoder respectively:

\begin{equation}
    G_\theta(f_t) = \mathbf{E}_{z_t \sim q_\theta(z_t|f_t)} [z_t],
\end{equation}

\begin{equation}
    G^{-1}_\theta(z_t) = \mathbf{E}_{f_t \sim p_\theta(f_t|z_t)} [f_t],
\end{equation}
where $p_\theta(f_t|z_t)$ is a likelihood function.

Convolutional neural networks \cite{colah-conv-2015, alexnet-2012, vgg-2014} are used to represent both likelihood function $p_\theta(f_t|z_t)$ and posterior distribution $q_\theta(z_t|f_t)$. The use of convolutional layers allows us to overcome both linearity and indifference to the spatial structure of tensors $f_t$. We use 3D convolutional layers with elementwise nonlinearity between them (Leaky Rectified Linear Unit -- LReLU~\cite{leakyrelu-2015}). Dimensionality reduction comes from the stride of the convolution window. 

\begin{figure}
    \centering
    \includegraphics[width=1\linewidth]{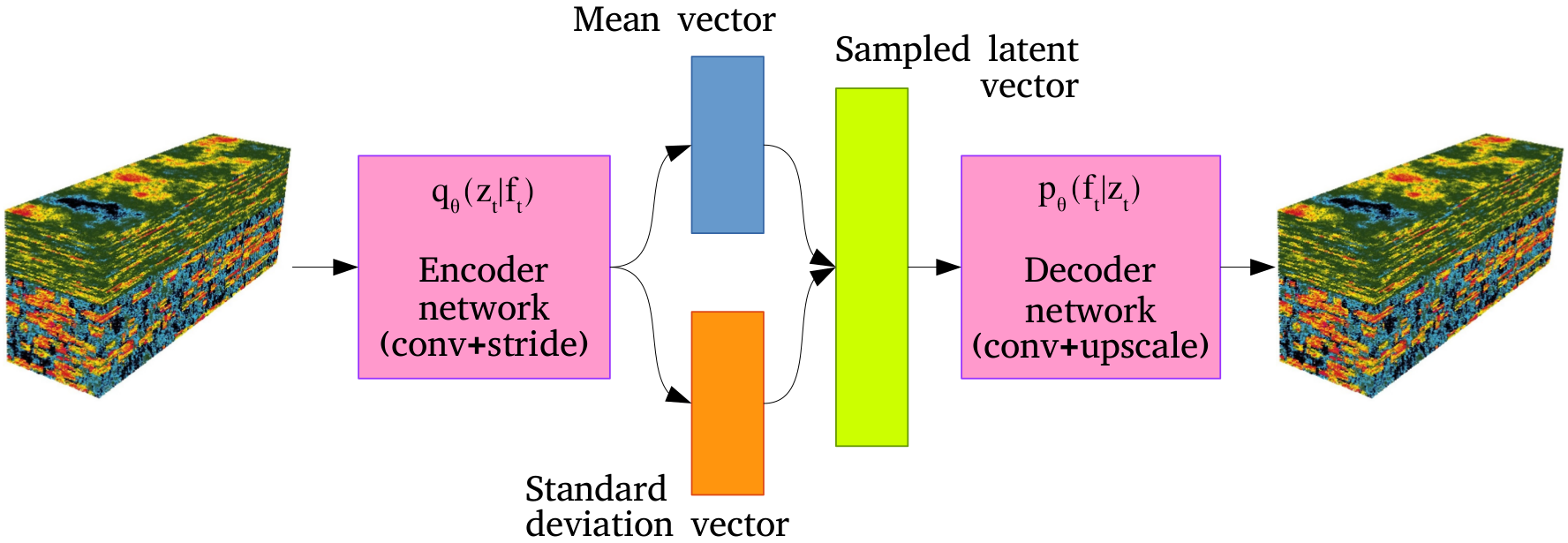}
    \caption{Graphical representation of convolutional Variational Autoencoder. Decoder network is composed of convolutional layers with stride, and encoder is composed of convolutional and upscaling (trilinear interpolation) layers.}
    \label{3d_conv}
\end{figure}

Convolutions exploit spatial invariance property that is also presented in FDHS engine: time derivatives of a property in a computational cell are dependent only on the parameters in the neighboring cells. Spatial invariance allows using less number of trainable parameters $\theta$ which helps us to overcome overfitting problem. 

Graphical representation of convolutional VAE is provided at Figure \ref{3d_conv}.

\subsection{Approximation of the latent dynamics}\label{dynamics-subsec}

Given a set of latent variables $z_t = G_\theta(f_t) \; \forall f_t \in \mathbf{D}$ we can now approximate the dynamics $F^{latent}$ with $\hat F^{latent}$ using standard ML techniques.

Most of ML models assume the input variable of a fixed dimensionality, which restricts the direct applicability of these models to the approximation of non-Markovian dynamics $F^{latent}$ since its inputs $z_{0:t}$ and $u_{0:t}$ are of variable length.
For this kind of models we may allow the approximation to be Markovian instead: $\hat z_{t+1} = \hat F^{latent}(\hat z_t, u_t, m) $.

Different possible choices of the latent dynamics model are provided below. We denote the set of model's parameters by $\phi$ and will fit them to minimize the average L2 loss in the latent space.

\textit{Linear Regression (LR)} model requires Markovian simplification and follows the equation:

\begin{equation}
    \hat z_{t+1} = W_z \hat z_t + W_u u_t + W_m m + b,
\end{equation}
where parameters $\phi=\{ W_z, W_u, W_m, b \}$ are obtained analytically.

\textit{Fully-Connected Neural Network (FCNN)} is another Markovian model able to approximate a much broader class of functions than LR. FCNN is composed of $L$ consecutive fully-connected layers with outputs $a^l, \; \forall l \in [1,\;\dots, L]$:

\begin{align}
    &a^l = \sigma(W^l a^{l-1} + b^l), \\
    &a^0 = concatenation(\hat z_t, u_t, m),\\
    &\hat z_{t+1} = a^L,
\end{align}
where $\sigma$ is an elementwise nonlinear function. Parameters $\phi = \bigcup_{l=1}^L \{W^l, b^l\}$ might be trained via one of gradient-based optimization procedures.

\textit{Recurrent Neural Network with Gated Recurrent Units (GRU RNN)}~\cite{gru-2014} compared to FCNN and LR models is a non-Markovian one. GRU RNN is a modification of a simple Recurrent Neural Network, and its internal structure is out of the scope of this work. For the sake of brevity, we provide here the equations for a simple RNN model. For $\forall t \in [0,\;\dots, T-1]$:

\begin{align}
    &h_{t+1} = \sigma_r(W_r h_t + W_z \hat z_t + W_u u_t + W_m m + b),\\
    &\hat z_{t+1} = \sigma_o(W_o h_{t+1} + b_o),
\end{align}
where $h$ is a hidden layer vector, $\sigma_o$ and $\sigma_r$ are nonlinear functions; parameters $\phi=\{ W_o, W_r, W_z, W_u, W_m, b_o, b \}$ are obtained from gradient-based optimization procedure.

We also tried to compare GRU RNN with closely related LSTM \cite{lstm-1997} model. In the experimental section, we showed almost the same accuracy of GRU and LSTM given the fact that LSTM has a bigger amount of trainable parameters (it is slower).

\subsection{Initial latent state approximation}

Each of described above latent dynamics models requires an initial latent state $\hat z_0$ to start the computations for a scenario.

From the equation (\ref{initial_states}) we know, that initial reservoir state $f_0$ is a function of just metadata vector $m$: $f_0 = F^{init}(m)$. Hence, $\hat z_0$ may be computed as follows:

$$ \hat z_0 = G_\theta(F^{init}(m)),$$
where $F^{init}$ is a known physics-based procedure aimed to calculate equilibristic initial reservoir state. 

However, this approach requires explicit calculation of the high-dimensional tensor $f_0 = F^{init}(m)$. Instead we may want to find the direct linear approximation for $\hat z_0$:
$$\hat z_0 \approx W_i m + b_i.$$

\section{Calculation of Well Production Rates}\label{well-prodrates-sec}

Besides the calculations of the dynamics, the metamodel is required to approximate the production rates function $R$. 

To calculate the production rates from the well we use the physics-based model without trainable parameters. We exploit the fact that there are no additional sources or sinks in the development unit except for production and injection wells. Hence, the amount of oil, water, and gas produced might be calculated from the mass balance as the difference between subsequent reservoir states. 

The volumetric amount of a fluid $\alpha$ at time $t$ can be calculated from the corresponding saturation field $s_t^\alpha$. Amounts of fluid from each computational cell should be brought to the same pressure $p_{ref}$ and then summed: 

\begin{equation}
    V^\alpha_t(p_{ref}) = V^{bulk} \phi(p_{ref})\sum_{i=1}^{n_x n_y n_z} s^\alpha_{ti} e^{(c_r - c_\alpha)(p_{ti}-p_{ref})}, 
\end{equation}
where pressure and saturation distributions $p_t$, $s^\alpha_t$ are taken from the approximated reservoir state $\hat f_t$, $V_{bulk}$ is a bulk volume of computational cell, $\phi$  is porosity, $c_r$ and $c_\alpha$ are rock and fluid compressibilities, and the index $i$ denotes the number of a computational cell.

The reference pressure $p_{ref}$ should be chosen from the interval where there is no mass transfer involved between the oil and the gas phases. For example, it might be selected to be equal to the initial pressure at the upper part of the reservoir.

Thus we can calculate the liquid's production rate at the reference pressure:

\begin{equation}
    r_t^\alpha(p_{ref}) = \frac{V^\alpha_t(p_{ref}) - V^\alpha_{t-1}(p_{ref})}{\Delta t} + q_t^\alpha(p_{ref}),
\end{equation}
where $q_t^\alpha$ is the injected amount of fluid $\alpha$, which is equal to zero for all fluids except for water.

We can now obtain the production rates at surface conditions as follows:

\begin{equation}\label{rates_surface}
    r_t^\alpha(p_{surface}) = \frac{r_t^\alpha(p_{ref})}{b_\alpha(p_{ref})},
\end{equation}
where $b_\alpha$ is a formation volume factor of the fluid $\alpha$.

To take into account the influence of gas, liberated from the oil, we use the following modification of (\ref{rates_surface}):

\begin{equation}
    r_t^g(p_{surface}) = \frac{r_t^g(p_{ref})}{b_g(p_{ref})} + \gamma r_t^o(p_{ref}),
\end{equation}
where $\gamma$ is the gas content.

\section{Summary}

The overall structure of the metamodel consists of two distinct parts: dynamics and production rates models. The dynamics model itself decomposes into the dimensionality reduction model and the latent dynamics model.

\begin{algorithm}[ht!]
\SetAlgoLined
\KwIn{Generative Model ($GM$) for metadata and control, \\ 
number of training scenarios $N$, \\
Base Model ($BM$)}
 Generate $\mathbf{m}, \mathbf{u} \sim GM $ for $N$ scenarios\;
 Compute $\mathbf{f}, \mathbf{r} \leftarrow BM(\mathbf{m}, \mathbf{u}) $\;
 Form dataset $\mathbf{D} \leftarrow (\mathbf{m}, \mathbf{u}, \mathbf{f}, \mathbf{r})$\;
 Train Autoencoder ($G_\theta, G_\theta^{-1}$) on $\mathbf{f} \in \mathbf{D}$\;
 Encode $\mathbf{z} = G_\theta(\mathbf{f})$\; 
 $\mathbf{D}^{latent} \leftarrow (\mathbf{m}, \mathbf{u}, \mathbf{z}, \mathbf{r})$\;
 Train Latent Dynamics $F^{latent}_\phi$ on $\mathbf{D}^{latent}$\;
 \Return ($G_\theta, G_\theta^{-1}, F^{latent}_\phi$)
 \caption{Generic training procedure for metamodels with the latent variable engine.}
 \label{training-alg}
\end{algorithm}

In the experimental section (Section \ref{experiments-sec}) we show the superiority of the dynamics model composed of VAE for dimensionality reduction and GRU RNN for latent dynamics forecasting against other investigated choices for these components of the dynamics model. 

Algorithm \ref{training-alg} shows the overall training procedure for the metamodel and is suitable for different choices of metamodel's components.

\section{Results}\label{experiments-sec}

To analyse the performance of proposed metamodelling approach, we generated a dataset (using the generative model described in \ref{gen-model-appendix}) and divided it into training $\mathbf{D}^{train}$ and validation $\mathbf{D}^{val}$ subsets, containing more than 5000 and 100 scenarios respectively. The dimensionality of the computational grid is chosen to be $(n_x, n_y, n_z) = (41, 60, 10)$ and time discretization step: $\Delta t = 14$ days.

The dataset contains scenarios of three types of possible development units, with either horizontal or vertical wells, different rock, and fluid properties, smooth or sharp applied control sequences $\mathbf{u}$. 

3D ground truth solutions $\mathbf{f}$ and production rates $\mathbf{r}$ were obtained from Rock Flow Dynamics tNavigator simulator~\cite{tnav} used as the Base Model. We turned on the options of the simulator responsible for gas liberation and oil evaporation. ``No flow'' boundary conditions and equilibristic initial conditions were used throughout the simulations. 

Following the Algorithm \ref{training-alg}, we performed the training for two variations of the dimensionality reduction model: VAE and PCA, and four variations of latent dynamics model: LR, FCNN, GRU and LSTM RNNs. Two dimensionalities of the latent space were tested: $d_z = 150$ and $d_z = 250$.

We used ADAM~\cite{adam-2014} optimizer to perform gradient-based optimization when needed. All the Neural Networks were implemented using standard NN modules from the PyTorch library~\cite{pytorch} for Python programming language. We used LReLU nonlinearity for all of the NNs, except the RNN modules where we use nonlinearities proposed in the original works~\cite{gru-2014, lstm-1997}. Batch normalization~\cite{batchnorm-2015} was used in all non-recurrent parts of NNs and layer normalization~\cite{layrnorm-2016} in recurrent ones.

It should be mentioned that the trained metamodel needs no additional training to be applied to a new development unit. The ideal case for us is to construct a dataset broad enough, that the metamodel can be trained just once and then applied to arbitrary reservoirs.

\begin{table}
    \centering
    \begin{tabular}{ |cc||c|c|c||c|c|c| }
     \hline
     \multicolumn{8}{|c|}{Mean relative error for $f_t$ in \% and its standard deviation} \\
     \hline\hline
     \multirow{2}{3.5em}{Model} & \multirow{2}{1em}{$d_z$} & \multicolumn{3}{c||}{Error $\pm$ std, \%} & \multicolumn{3}{c|}{RMSE} \\
     & & $p$ & $s_{oil}$ & $s_{wat}$ & $p$ (bar) & $s_{oil}$ & $s_{wat}$ \\
     \hline\hline
     Linear+PCA & $150$ & $16.5\pm10.8$ & $59.5\pm15.4$ & $38.5\pm7.7$ & 32.88 & 0.26 & 0.27 \\
     \hline
     Linear+VAE & $150$ & $28.7\pm20.2$ & $35.4\pm8.5$ & $22.4\pm8.8$ & 63.42 & 0.16 & 0.16 \\
     \hline
     Linear+VAE & $250$ & $11.1\pm8.7$ & $22.7\pm12.4$ & $12.8\pm6.5$ & 21.42 & 0.10 & 0.09 \\
     \hline
     GRU+PCA & $150$ & $43.9\pm30.1$ & $67.3\pm5.6$ & $42.4\pm9.4$ & 90.33 & 0.30 & 0.30 \\
     \hline
     GRU+VAE & $150$ & $\boldsymbol{8.0\pm5.1}$ & $\boldsymbol{11.4\pm6.2}$ & $\boldsymbol{6.7\pm3.2}$ & 19.13 & \textbf{0.05} & \textbf{0.05} \\
     \hline
     GRU+VAE & $250$ & $9.5\pm6.5$ & $14.7\pm8.5$ & $8.6\pm4.2$ & 21.37 & 0.06 & 0.06 \\
     \hline
     LSTM+VAE & $150$ & $8.9\pm6.9$ & $12.8\pm8.7$ & $7.5\pm4.6$ & \textbf{18.72} & 0.06 & 0.05 \\
     \hline
     
    \end{tabular}
    \caption{Prediction error between forecasted and ground truth reservoir states $f_t$ for different metamodels. All the values are averaged throughout the validation dataset.}
    \label{dynamics-table}
\end{table}

\begin{figure}[ht!]
    \centering
    \subfloat[pore pressure]{\includegraphics[width=.5\linewidth]{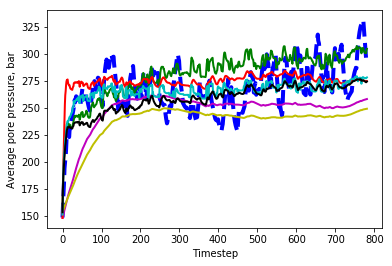}}
    \subfloat[oil saturation]{\includegraphics[width=.5\linewidth]{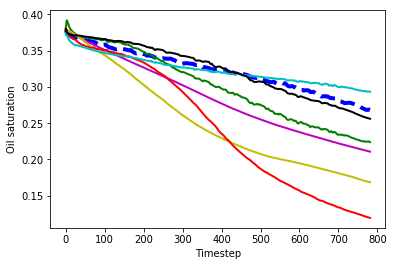}}
    \\
    \subfloat[water saturation]{\includegraphics[width=.5\linewidth]{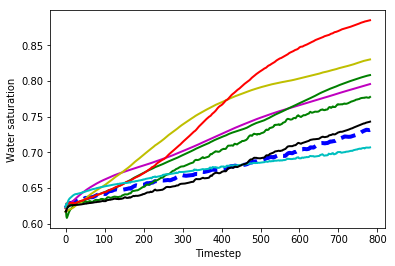}}
    \subfloat{\includegraphics[width=.25\linewidth]{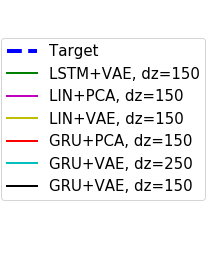}}
    \caption{Mean values of spatial parameters calculated for a single scenario and averaged across all computational cells versus time: target and predicted by metamodels.}
    \label{mean-vs-time}
\end{figure}

\begin{figure}[ht!]
    \centering
    \subfloat{\includegraphics[width=.3\linewidth]{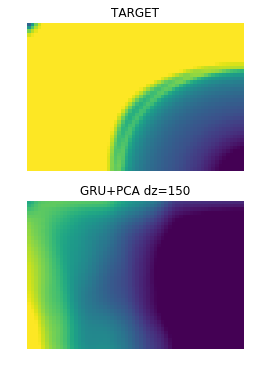}}
    \subfloat{\includegraphics[width=.3\linewidth]{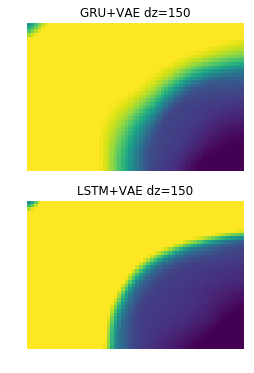}}
    \subfloat{\includegraphics[width=.3\linewidth]{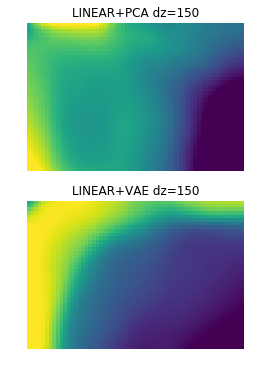}}
    \subfloat{\includegraphics[width=.1\linewidth]{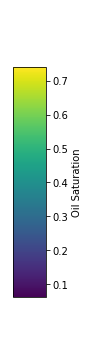}}
    \caption{Horizontal slices of the oil saturation tensor $s^o_t$ at the medium layer at time $t=15$ years: target and predicted by metamodels.}
    \label{oslices}
\end{figure}
\textit{Dynamics} part of the metamodel was evaluated in the sence of relative deviation and Rooted Mean Squared Error (RMSE) between metamodel's forecast and the ground truth for all scenarios from the validation dataset $\mathbf{D}^{val}$. The results of the analysis are provided in the Table \ref{dynamics-table} for different choices of metamodel's components. Bold values denotes the best obtained approximation accuracy.

The FCNN model is not provided in the table due to its inability to output long predictions. The training scheme for FCNN is aimed to minimize the error of one-step prediction. This leads FCNN to the extreme instability when NN gets its outputs as inputs. Interestingly, that this effect does not appear in the LR model, which is structured similarly to FCNN model, due to the simplicity of linear regression.

Visual analysis of the dynamics forecasting performance for the 3D flow problem was made using the plots of mean pressure and fluid saturations versus time (Figure \ref{mean-vs-time}) and with the horizontal slices of pressure and saturation tensors at a fixed moment of time (Figure \ref{oslices}). Figures \ref{mean-vs-time} and \ref{oslices} show the results for the randomly chosen validation scenario with sharp applied control, staggered line drive allocation scheme, and vertical wells.

\begin{figure}[ht!]
    \centering
    \subfloat[allocation scheme]{\includegraphics[width=.33\linewidth]{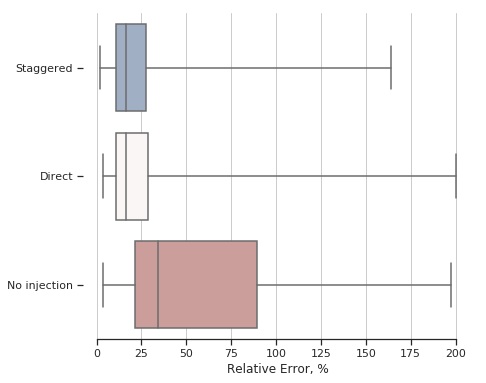}}
    \subfloat[applied control]{\includegraphics[width=.33\linewidth]{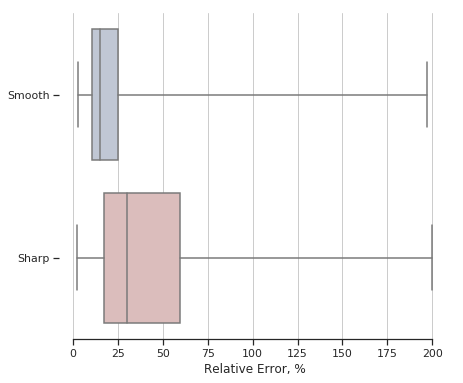}}
    \subfloat[well orientation]{\includegraphics[width=.33\linewidth]{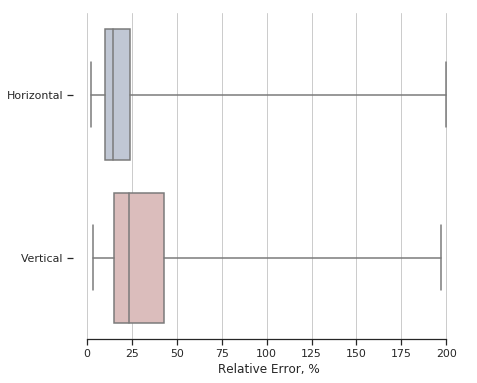}}
    \caption{The distribution of production rates prediction error between different groups of validation scenarios. VAE + GRU RNN, $d_z = 150$ metamodel was used to make predictions.}
    \label{rates-histogram}
\end{figure}

\begin{figure}[ht!]
    \centering
    \subfloat[water production rate]{\includegraphics[width=.33\linewidth]{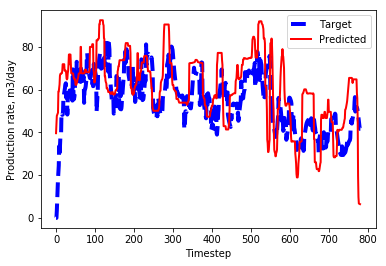}}
    \subfloat[oil production rate]{\includegraphics[width=.33\linewidth]{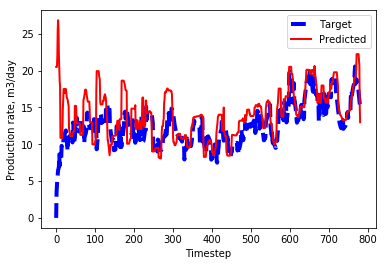}}
    \subfloat[gas production rate]{\includegraphics[width=.33\linewidth]{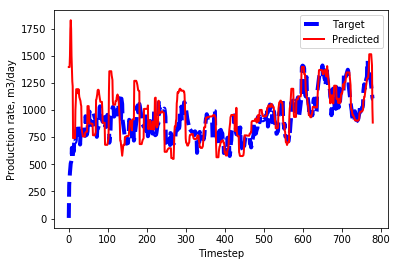}}
    \caption{Ground truth and predicted production rates for the randomly chosen scenario. VAE + GRU RNN, $d_z = 150$ metamodel was used to make predictions.}
    \label{rates-prediction}
\end{figure}

The accuracy of the procedure for computing \textit{production rates} was tested on the best dynamics metamodel (VAE + GRU RNN, $d_z = 150$) since it is strongly dependent on the dynamics accuracy. Mean relative deviation between predicted and ground truth rates for this model is provided on the Figure \ref{rates-histogram} where different groups of scenarios were analyzed separately. Visual check of the performance can be made on the plots of predicted daily production rates for randomly chosen validation scenario (Figure \ref{rates-prediction}) -- here we used the same scenario that was used in the dynamics visualization.




On average, the total on-wall time consumed by tNavigator to produce solutions for 30 years of simulation is 120 seconds. Computations were divided between 30 CPU cores. On the other hand, the metamodel consumes 10 seconds for the same simulation, if the GPU is used. Even in the case when the CPU is used, it consumes only 70 seconds. Though, convolutional layers are not adapted to be used with CPU. So, the metamodel is about x12 faster than the Base Model.

\section{Discussion}

We have developed the novel approach for prediction of reservoir performance enabling a new way of assessing the possible reservoir development scenario. Being much faster than a conventional reservoir modeling tool, the metamodel allows overlooking many more scenarios over a given time interval. 

The approach is a pioneering one regarding its ability to output not only the flow rates at the wellheads, but also the actual dynamics of 3D fields of pressure, and fluid saturations over the computational cells of a conventional hydrodynamic model of a reservoir.

The approach itself is also somewhat unique because of its high-level structure containing methods of reduced order modelling (ROM) and the data-driven forecasting (DDF) tools. This high-level structure presents a very general framework for further enhancements as both ROM and the DDF methods are being developed these days very rapidly. We have tried just several ROM and DDF approaches and defined the ones demonstrating the best results. 

There is no doubt that both building blocks of our workflow will progress in terms of accuracy and computational efficiency in training and especially prediction modes. 

\section{Further Work}

We plan to continue our research in three principal directions. First is further experimenting with DDF and ROM architectures to increase the accuracy of the forecast and ensure the higher computational efficiency, as well as to try to combine data, generated by FDHS with different fidelities, when training a metamodel (see also \cite{MFGP2015,MFGP2017}). Second is about going a scale up and try to adapt the metamodels for the full-scale reservoirs and oilfields containing many development units. Later will lead to more useful tools for applied reservoir engineering, including field development optimization and history matching. The third is the expansion of the metamodel's operational envelope to the more complex fluid systems and their PVT representations and the low permeable reservoirs. The low permeable direction is likely to be supported by a training set generated with an approach presented in \cite{bezyan2019novel}.  

\bibliographystyle{unsrt}
\bibliography{\jobname}
\appendix

\section{Scenario metadata in compressed vector form}\label{metadata-appendix}

Variable $m$, defined in Section \ref{metamodelling-sec} and mentioned as metadata of the development unit, represents physical properties of the rock and fluids, and geometrical characteristics of the development unit and corresponding wells.

We use the metadata as an input of ML models, and most of them require inputs in a vectorised form with fixed dimensionality. Moreover, this dimensionality should be small to prevent the overfitting. 

Some of the physical properties can be represented as scalars and used as is, being stacked into a vector. These are rock and water compressibilities, fluids densities, water viscosity, development unit size in each of three dimensions, initial water-oil and gas-oil contact depths, the depth of the top of the reservoir, initial pore pressure, saturation pressure and gas content.

Categorical properties, namely well allocation scheme and spatial orientation of wells, can be presented via one-hot encoding. The exact position of either vertical or horizontal well can be described by two values: the depth of the beginning of the perforation interval, the length of the perforation interval. We assume wells to have just one perforation interval.

The complicated part is to represent properties defined for each computational cell (porosity, absolute permeability) and table-defined functions (PVT tables, tables of relative permeabilities).

We exploit the restrictions imposed on the development unit to compress large fields such as porosity and permeability fields. We assume homogeneous rock in the sense of porosity, and layerwise-nonhomogeneous rock in the sense of permeability. So the porosity can be represented as a scalar, and permeability field forms the vector of length $n_z$ with values for each layer of the reservoir.

The problem with table-defined functions is that they have a non-fixed number of points in their tables. The problem can be resolved by interpolating these functions onto a predefined domain.

Such representation fixes the dimensionality but is rather high-dimensional. We compress the dimensionality using the PCA method. 

Actually, the restrictions on the homogeneity of the development unit might be overcome using the same dimensionality reduction idea (possible with the neural network instead of PCA). Though, this approach requires construction of the generative model for the whole heterogeneous permeability field, which is a complicated task by itself.

\section{Generative model of realistic metadata and control variables}\label{gen-model-appendix}

Generation of synthetic but realistic oil-fields is a very complex task by itself. We restricted ourselves on the generation of just pieces of an oil-field (development units) with the layered structure.

But still, the detailed description of the process is too large to be shown in the paper. The generative model should provide us with the randomised samples of the reservoir geometry, porosity and absolute permeability fields, initial conditions, PVT and relative permeability tables, exact positions of the wells, and so on.

Here we present the overall ideas used to obtain the generative model:

\begin{itemize}
    \item Collect a large set of values of the interest from a variety of real oil-fields.
    \item Divide all the properties into interindependent groups. 
    \item Visualize joint empirical distributions for each group.
    \item For each group, fit parameters of chosen parametric distribution to match the empirical data using the max-likelihood method. We used following parametric families of distributions: Gaussian, log-Gaussian, Uniform.
\end{itemize}

\begin{figure}[ht!]
    \centering
    \includegraphics[width=.5\linewidth]{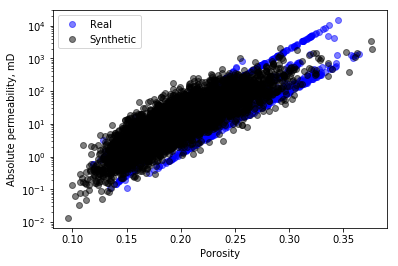}
    \caption{Joint distribution over log-permeability and porosity values. Plot shows points from the dataset of the real values (blue) and points drawed from the fitted Gaussian distribution (black).}
    \label{poroperm}
\end{figure}

For example, we grouped values of porosity and absolute permeability and modelled them jointly. The form of the empirical distribution is similar to two dimensional Gaussian if the permeabilities are logarithmized. Samples from the resulting distribution are shown in Figure \ref{poroperm} together with the points from the dataset.

Table-defined functions are generated from the point estimates paired with standard correlation functions, such as Corey, Al-Marhoun, Khan et al., Vasquez and Beggs correlations~\cite{correlation-2014}.

\begin{figure}[ht!]
    \centering
    \subfloat[exponentional]{\includegraphics[width=.4\linewidth]{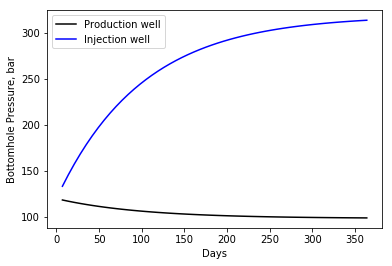}}
    \subfloat[sharp]{\includegraphics[width=.4\linewidth]{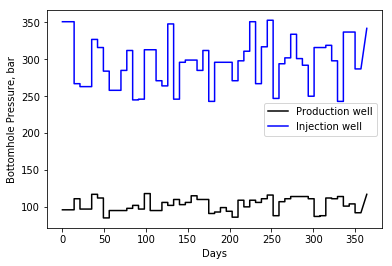}}
    \caption{Samples of two types of applied control variables: exponentional decay and sharp changes. Time-series of bottomhole pressure for injection and production wells.}
\label{control_samples}
\end{figure}

For the time-series (bottomhole pressures and injection rates) we used three possible forms: constant, exponential, sharp (see Fig.~\ref{control_samples}).
The form is sampled randomly, and actual values are obtained from the Uniform distribution.  

\end{document}